\documentclass[conference]{IEEEtran}
\IEEEoverridecommandlockouts
\usepackage{cite}
\usepackage{amsmath,amssymb,amsfonts}
\usepackage{algorithmic}
\usepackage{graphicx}
\usepackage{textcomp}
\usepackage{xcolor}
\usepackage{multirow}
\usepackage{colortbl}  
\usepackage{array}

\usepackage{stfloats}
\usepackage{url}
\usepackage{verbatim}
\usepackage{cite}
\usepackage{color}
\usepackage{booktabs}
\usepackage{utfsym}
\usepackage{soul}

\definecolor{gg}{HTML}{e2f0cb}

\def\BibTeX{{\rm B\kern-.05em{\sc i\kern-.025em b}\kern-.08em
    T\kern-.1667em\lower.7ex\hbox{E}\kern-.125emX}}
\begin{document}

\title{Enhancing Robustness of Implicit Neural Representations Against Weight Perturbations
\thanks{*: Equal contributions. \textsuperscript{\dag}: Corresponding authors: Zhengwu Liu and Ngai Wong \{zwliu, nwong@eee.hku.hk\}. This work was supported in part by the Theme-based Research Scheme (TRS) project T45-701/22-R, National Natural Science Foundation of China (62404187) and the General Research Fund (GRF) Project 17203224, of the Research Grants Council (RGC), Hong Kong SAR.}
}

\author{
\IEEEauthorblockN{Wenyong Zhou*}
\IEEEauthorblockA{\textit{Department of EEE} \\
\textit{The University of Hong Kong}\\
Hong Kong SAR}\\

\IEEEauthorblockN{Taiqiang Wu}
\IEEEauthorblockA{\textit{Department of EEE} \\
\textit{The University of Hong Kong}\\
Hong Kong SAR}

\and

\IEEEauthorblockN{Yuxin Cheng*}
\IEEEauthorblockA{\textit{Department of EEE} \\
\textit{The University of Hong Kong}\\
Hong Kong SAR}\\

\IEEEauthorblockN{Chen Zhang}
\IEEEauthorblockA{\textit{Department of EEE} \\
\textit{The University of Hong Kong}\\
Hong Kong SAR}

\and

\IEEEauthorblockN{Zhengwu Liu\textsuperscript{\dag}}
\IEEEauthorblockA{\textit{Department of EEE} \\
\textit{The University of Hong Kong}\\
Hong Kong SAR}\\

\IEEEauthorblockN{Ngai Wong\textsuperscript{\dag}}
\IEEEauthorblockA{\textit{Department of EEE} \\
\textit{The University of Hong Kong}\\
Hong Kong SAR}
}

\maketitle

\begin{abstract}
Implicit Neural Representations (INRs) encode discrete signals in a continuous manner using neural networks, demonstrating significant value across various multimedia applications. However, the vulnerability of INRs presents a critical challenge for their real-world deployments, as the network weights might be subjected to unavoidable perturbations. In this work, we investigate the robustness of INRs for the first time and find that even minor perturbations can lead to substantial performance degradation in the quality of signal reconstruction. To mitigate this issue, we formulate the robustness problem in INRs by minimizing the difference between loss with and without weight perturbations. Furthermore, we derive a novel robust loss function to regulate the gradient of the reconstruction loss with respect to weights, thereby enhancing the robustness. Extensive experiments on reconstruction tasks across multiple modalities demonstrate that our method achieves up to a 7.5~dB improvement in peak signal-to-noise ratio (PSNR) values compared to original INRs under noisy conditions.
\end{abstract}


%%%%%%%%%%%%%%%%%%%%%%%%%%%%%%%%%%%%%%
%             Introduction           %
%%%%%%%%%%%%%%%%%%%%%%%%%%%%%%%%%%%%%%
\section{Introduction}
\label{sec:introduction}
Implicit Neural Representations (INRs) provide a flexible approach to encoding various signals, including images, audio, and video~\cite{mildenhall2020nerf,sitzmann2020implicit, tancik2020fourier, chen2022videoinr}. INRs model signals as continuous functions parameterized by neural networks, typically implementing Multi-Layer Perceptrons (MLPs)~\cite{li2023regularize}. For example, when modeling a two-dimensional grayscale image, INRs map pixel coordinates to their corresponding grayscale values, enabling high-quality signal reconstruction while supporting signal interpolation and super-resolution. Due to these capabilities, INRs have gained significant traction in various applications~\cite{park2019deepsdf,sitzmann2019srns, grattarola2022generalised}.

%%%%%%%%%%%%%%%%%%
\begin{figure}[!t]
\centering
\includegraphics[scale=0.32]{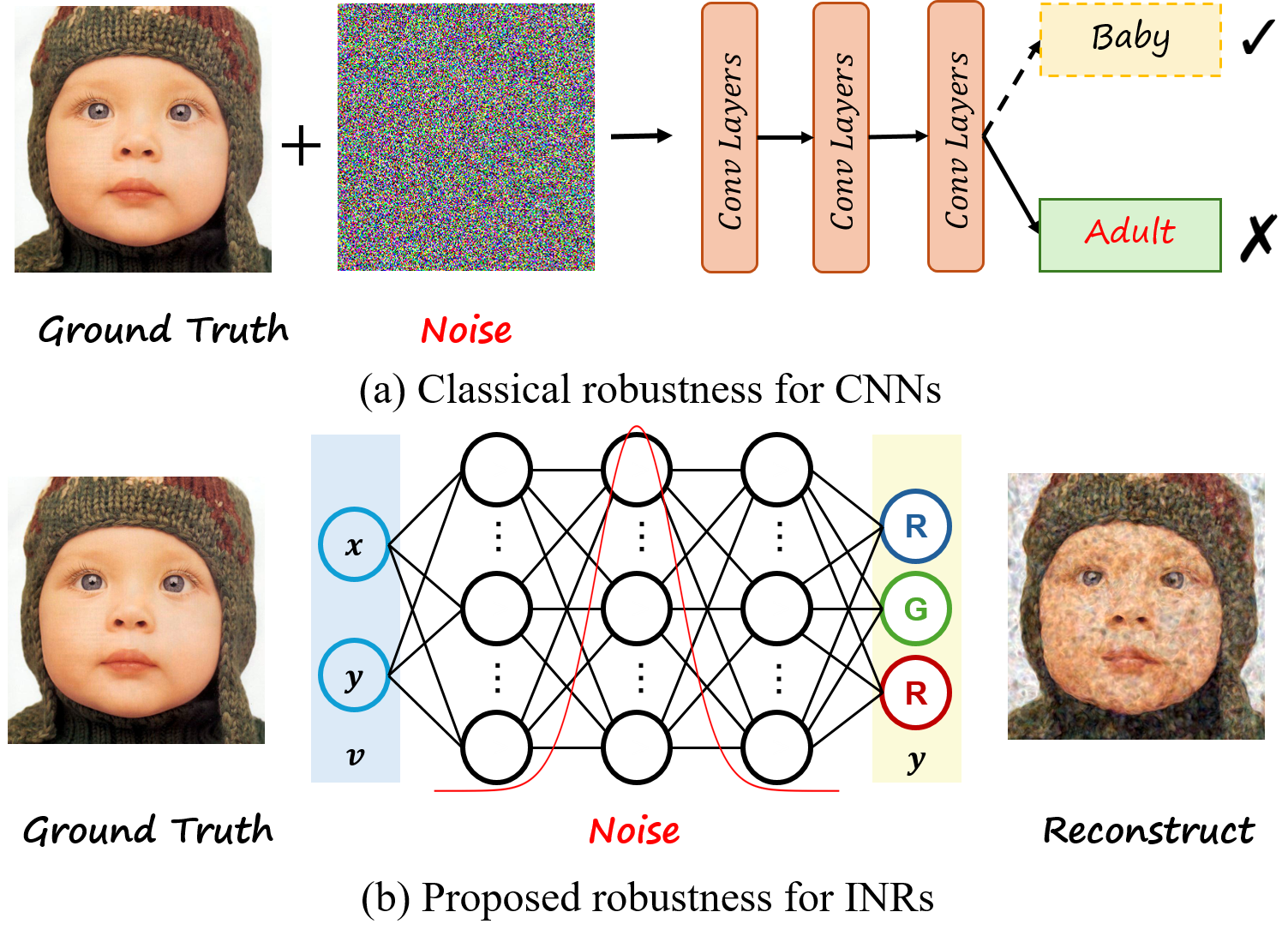}
\vspace{-0.2cm}
\caption{Comparison of robustness between classical CNNs and INRs: (a) In classical CNNs, noise typically corrupts input images, leading to misclassification or errors in the task. (b) In INRs, noise perturbs the model weights, resulting in degraded reconstruction performance.} 
\label{fig:motivation}
\vspace{-0.2cm}
\end{figure}
%%%%%%%%%%%%%%%%%

Previous works have primarily focused on improving the performance and efficiency of INRs. Various complex activation functions, including sinusoidal, Gaussian, and wavelet functions, have been explored to enhance the high-frequency representation capabilities of neural networks in INRs, enabling them to capture finer signal details~\cite{ramasinghe2022beyond}. To accelerate the training process,~\cite{seo2024search} proposed a simple random permutation of pixel locations, while~\cite{zhang2024nonparametric} introduced an efficient data sampling framework by formulating INR training as a nonparametric teaching problem. However, a critical challenge overlooked by existing work is the robustness of INRs.

The robustness challenge in INRs is fundamentally distinct from that in Convolutional Neural Networks (CNNs). In CNNs, robustness primarily targets resistance to adversarial samples that subtly perturb input images with minimal content changes, as depicted in Figure~\ref{fig:motivation}~\cite{boopathy2019cnn, zhang2023robust, benbarka2022seeing, lin2019towards}. In contrast, INR robustness concerns shift from input data to model weights, as the signal is encoded directly in the network parameters. Sources of weight perturbations include model transmission errors, compression inaccuracies, and hardware constraints in devices like compute-in-memory (CIM) systems\cite{sattler2019robust, han2015deep, chen2021distributed, chen2020reram}. Even minimal perturbations can significantly impair INR performance due to the sensitivity of reconstruction tasks~\cite{grattarola2022generalised, lipman2021phase}. 
Furthermore, traditional robustness-enhancing methods like adversarial training and data augmentation are less effective for INRs~\cite{dey2017regularizing, saragadam2023wire}. Adversarial training typically generates perturbed input samples to improve model robustness, but in INRs, the input coordinates remain fixed and unperturbed since they serve as query points for signal reconstruction. Data augmentation methods that modify input data (e.g., rotation, scaling, or adding noise) do not align with the INR paradigm, where the challenge lies in weight perturbations rather than input variations.

To address this challenge, we present the first systematic investigation of INR vulnerability to weight perturbations and propose a robust loss function to enhance model resilience. Our main contributions are threefold:
\begin{itemize} 
	\item We present a systematic investigation of the robustness of INRs under weight perturbation and find that even a tiny perturbation can lead to significant degradation in reconstruction quality.
	\item We formulate the robustness problem in the INR context by minimizing the reconstruction distance between clean and perturbed weights. Based on the formulation, we derive a novel robust loss function to enhance the robustness of INR.
	\item Extensive experiments on reconstruction tasks across multiple modalities demonstrate that our method yields improvements of up to 2.5 dB in PSNR values compared to the baseline INR under various noise conditions.    
\end{itemize}

%%%%%%%%%%%%%%%%%%%%%%%%%%%%%%%%%%%%%%
%             Methodology            %
%%%%%%%%%%%%%%%%%%%%%%%%%%%%%%%%%%%%%%
\section{Methodology}
\label{sec:method}
This section begins by demonstrating the vulnerability of INRs in the image reconstruction task. To address this issue, we formulate the problem and propose a gradient-based loss function to regularize the gradient of the reconstruction loss with respect to the model weights.
%%%%%%%%%%%%%%%%%%
\subsection{Vulnerability of INRs for Reconstruction Tasks}
\begin{figure}[!t]
\centering
\includegraphics[scale=0.33]{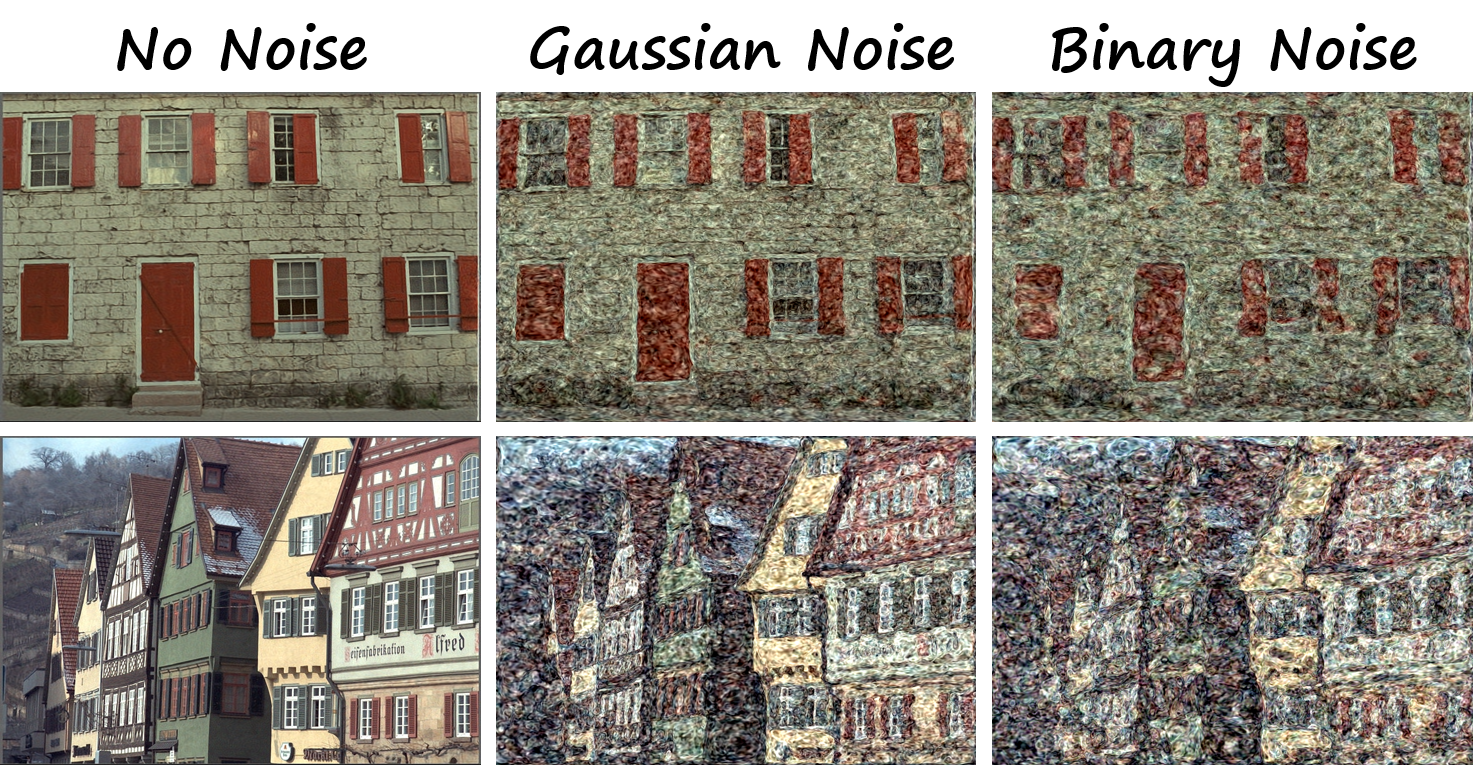}
% \vspace{-20pt}
\caption{The reconstruction of INRs with and without weight perturbations on {\tt Kodak01}, {\tt Kodak08} images under Gaussian and binary noises.} 
\label{fig:observation}
\vspace{-0.4cm}
\end{figure}

%%%%%%%%%%%%%%%%%%
Given the differences in robustness between CNNs and INRs, we first explore the impact of weight perturbations on INRs in the image reconstruction task, as visualized in Figure~\ref{fig:observation}. In the figure, the left column shows the original images reconstructed by the INR model without any noise. These images are clear and retain the original details, demonstrating the model's ability to accurately capture and represent the input data. The middle and right columns introduce Gaussian and binary noise into the model weights, resulting in a noticeable decline in image quality. While the images still maintain basic structure, the added noise causes significant blurring and distortion, underscoring the model's limited robustness to weight perturbations. It should be noted that the noise applied was only 1\% of the true value of the model weights, or only 1\% of the parameters were masked to zero. While this typically results in negligible performance degradation in classification tasks for CNNs, it leads to severe degradation in INRs.

%%%%%%%%%%%%%%%%%%

%%%%%%%%%%%%%%%%%%%%%%%%%%%%%%%%%%%%
\subsection{Gradient-Based Loss for Improving INR Robustness}
The previous section demonstrates the difference in noise impact on classical CNNs and INRs. To obtain a better understanding, we formulate the robustness problem in this subsection and then propose our gradient-based loss function. 

Suppose the continuous signal that we want to reconstruct is a function $f: \mathbb{R}^n \rightarrow \mathbb{R}^m$, where $n$ is the dimensionality of the input space (e.g., coordinates for an image) and $m$ is the dimensionality of the output space (e.g., 3 for RGB values). The INR model $\hat{f}_\theta$ is a MLP parameterized by $\theta$, which approximates $f$. Given a set of discrete samples $\{(\mathbf{x}_i, \mathbf{y}_i)\}_{i=1}^N$, where $\mathbf{x}_i \in \mathbb{R}^n$ are the input coordinates and $\mathbf{y}_i = f(\mathbf{x}_i)$ are the corresponding output values, the goal of the INR model is to learn the parameters $\theta$ to minimize the reconstruction loss:
%%%%%%%%%%%%%%%%%%%
\begin{equation} 
\mathcal{L}(\theta) = \frac{1}{N} \sum_{i=1}^{N} \left\| \hat{f}_\theta(\mathbf{x}_i) - \mathbf{y}_i \right\|^2_2
\end{equation}
%%%%%%%%%%%%%%%%%%%
When model weights $\theta$ are perturbed by some slight noise $\Delta \theta$, the new set of parameters $\theta + \Delta \theta$ will create a new image whose reconstruction loss is:
%%%%%%%%%%%%%%%%%%%
\begin{equation} 
\mathcal{L}(\theta + \Delta \theta) = \frac{1}{N} \sum_{i=1}^{N} \left\| \hat{f}_{\theta + \Delta \theta}(\mathbf{x}_i) - \mathbf{y}_i \right\|^2_2
\end{equation}
%%%%%%%%%%%%%%%%%%%
Although minimizing Eq. 2 yields the best reconstruction results, the presence of unknown noise makes it impossible to directly optimize the reconstruction loss under noisy conditions. Therefore, optimizing a proxy for the noisy reconstruction loss can help guide the model towards a more robust state. Since we want the model's output under noise to be as close as possible to the output in the absence of noise, minimizing the difference between the noisy reconstruction loss and the original reconstruction loss could serve as an effective proxy target. In other words, we aim to achieve $\mathcal{L}(\theta + \Delta \theta) \approx \mathcal{L}(\theta)$, or more precisely:
%%%%%%%%%%%%%%%%%%%
\begin{equation}
\min_\theta ||\mathcal{L}(\theta + \Delta \theta) - \mathcal{L}(\theta)||
\end{equation}
%%%%%%%%%%%%%%%%%%%
Since the noise $\Delta \theta$ is a small perturbation around the parameters $\theta$, it is possible for us to perform a first-order Taylor expansion around $\theta$, which is given by:
%%%%%%%%%%%%%%%%%%%
\begin{equation}
\mathcal{L}(\theta + \Delta \theta) = \mathcal{L}(\theta) + \nabla_\theta \mathcal{L}(\theta) \cdot \Delta \theta + o(\Delta \theta)
\end{equation}
%%%%%%%%%%%%%%%%%%%
where $\nabla_\theta \mathcal{L}(\theta)$ is the gradient of the loss function with respect to $\theta$, and $o(||\Delta \theta||)$ represents higher-order terms that are of smaller order compared to $\Delta \theta$. Omitting the higher-order terms, we could get:
%%%%%%%%%%%%%%%%%%
\begin{equation}
\begin{split}
    ||\mathcal{L}(\theta+\Delta \theta) - \mathcal{L}(\theta)|| &\approx ||\nabla_\theta \mathcal{L}(\theta) \cdot \Delta \theta|| \leq ||\nabla_\theta \mathcal{L}(\theta)|| \cdot ||\Delta \theta||
\end{split}
\end{equation}
%%%%%%%%%%%%%%%%%%
Since the noise $\Delta \theta$ is unknown and small, it is important to regularize the gradient term $\nabla_\theta \mathcal{L}(\theta)$ to minimize the difference between the noisy reconstruction loss and the
original reconstruction loss. Therefore, we introduce a gradient-based robust loss function $\mathcal{L}_{\text{robust}}(\theta)$:
%%%%%%%%%%%%%%%%%%
\begin{equation}
\mathcal{L}_{\text{robust}}(\theta) = \mathcal{L}(\theta) + \lambda \cdot  ||\nabla_\theta \mathcal{L}(\theta)||
\end{equation}
%%%%%%%%%%%%%%%%%%
where $\lambda$ is a hyperparameter that controls the trade-off between the original loss and the robustness penalty. 

During the backpropogation, the gradient of the norm of gradients with respect to \(\theta\) is derived as:
\begin{align}
\nabla_\theta \|\nabla_\theta \mathcal{L}(\theta)\| &= \nabla_\theta \sqrt{\sum_{i=1}^d \left( \frac{\partial \mathcal{L}}{\partial \theta_i} \right)^2} \\
&=\nabla_\theta \sqrt{\nabla_\theta \mathcal{L}(\theta) \cdot \nabla_\theta \mathcal{L}(\theta)}\\
&=\frac{\nabla_\theta \mathcal{L}(\theta)}{\|\nabla_\theta \mathcal{L}(\theta)\|}
\end{align}
The equation shows that the gradient of the norm is the original gradient vector scaled by the reciprocal of its norm. This computation only involves first-order derivatives of \(\mathcal{L}(\theta)\), which are typically computed during backpropagation, making it efficient.

%%%%%%%%%%%%%%%%%%%%%%%%%%%%%%%%%%%%%%
%             Experiments            %
%%%%%%%%%%%%%%%%%%%%%%%%%%%%%%%%%%%%%%
\section{Experiments}
\label{sec:experiment}
In this section, we compare our gradient-based robust loss function with previously proposed robust loss functions on signal reconstruction tasks across multiple modalities to demonstrate the effectiveness of our method.
\subsection{Experiment Setup}
Given the novel focus of our study, comparable methods within our specific domain are scarce. Consequently, we have adapted two established loss functions from CNN robustness studies for comparison: the $L_1$ loss~\cite{chensparsity} and the Lipschitz loss~\cite{lindefensive}. Additionally, we include a comparison with noise-aware training~\cite{noisetrain}, which involves injecting noise into the weights during training to simulate realistic noise conditions. All experiments are conducted on a single NVIDIA RTX 3090 GPU.

\subsection{Experiment Results}
\noindent \textbf{Main results.} We take images from the Kodak dataset as reconstruction images in this experiment and set $\sigma=1e^{-3}$ for the Gaussian noise condition and $p=1e^{-3}$ for binary noise. The result is shown in Table~\ref{tab:result}. For Gaussian noise, our method outperforms the other methods by yielding improvements in PSNR values by up to 2.53 dB, 3.01 dB, and 2.16 dB on different images, respectively. Under binary noise, our method again shows consistent superior performance by improving the PSNR values by up to 5.16 dB, 2.88 dB, and 5.30 dB for different images, respectively.

%%%%%%%%%%%%%%%%%%%%%%
\begin{table}[!t]
\renewcommand{\arraystretch}{1.2}
\caption{Comparison between our method and other robustness methods when reconstructing images in the Kodak dataset under various noises.}
\centering
\begin{tabular}{lccc}
\toprule
\multirow{2}{*}{\textbf{Methods}} & \multicolumn{3}{c}{\textbf{PSNR (dB)}} \\  % Changed 4 columns to 3 columns
\cmidrule{2-4} 
& {\tt Kodak01} & {\tt Kodak08}  & {\tt Kodak24}\\
\midrule
\multicolumn{4}{c}{\textit{Gaussian Noise ($\sigma=0.001$)}} \\
\midrule
MSE Loss                                            & 15.73       & 12.42             & 14.96  \\
\quad + $L_1$ Loss~\cite{chensparsity}              & 15.41       & 11.88             & 14.88  \\
\quad + Lipschitz Loss~\cite{lindefensive}          & 15.11       & 11.47             & 14.55 \\
\quad + Noise-aware Training~\cite{noisetrain}      & 15.26       & 11.57             & 14.62 \\
\rowcolor{gg} \textbf{Ours}                         & 17.64       & 14.48             & 16.71 \\
\midrule
\multicolumn{4}{c}{\textit{Binary Noise ($p=0.001$)}} \\
\midrule
MSE Loss                                            & 18.57       & 13.48             & 17.29  \\
\quad + $L_1$ Loss~\cite{chensparsity}              & 18.70       & 13.51             & 17.32  \\
\quad + Lipschitz Loss~\cite{lindefensive}          & 18.41       & 13.60             & 17.49 \\
\quad + Noise-aware Training~\cite{noisetrain}      & 17.32       & 13.50             & 17.08 \\
\rowcolor{gg} \textbf{Ours}                         & 22.48       & 16.36             & 22.38  \\
\bottomrule
\end{tabular}
\label{tab:result}
\vspace{-0.4cm}
\end{table}

%%%%%%%%%%%%%%%%%%%%%%
%%%%%%%%%%%%%%%%%%%%%%
\begin{figure}[!t]
\centering
\includegraphics[scale=0.32]{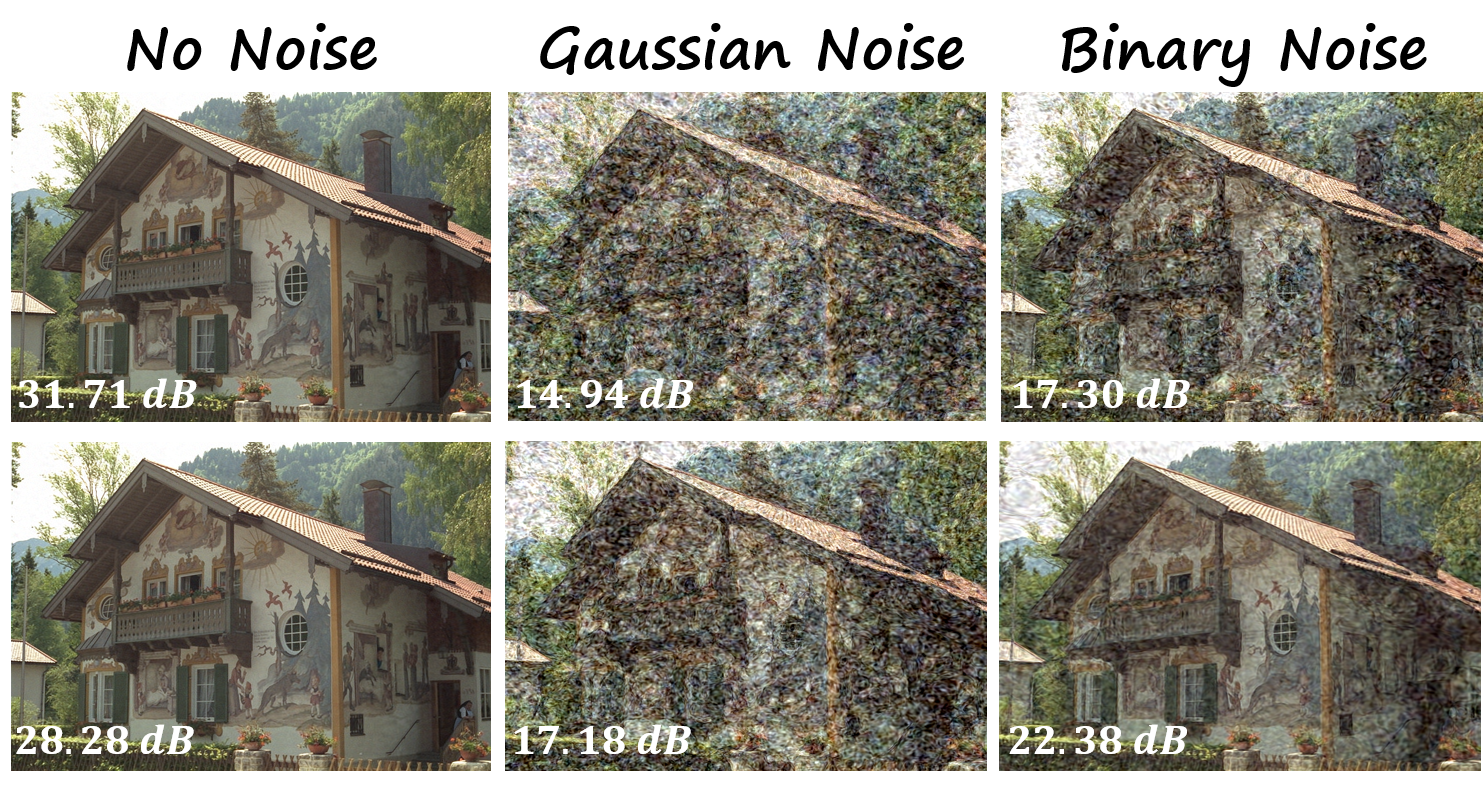}
\vspace{-5pt}
\caption{The reconstruction results between baseline SIREN (first row) and our method (second row) under noise-free, Gaussian noise and binary noise condition on {\tt Kodak24} images.} 
\label{fig:rgb_image}
\vspace{-0.4cm}
\end{figure}
%%%%%%%%%%%%%%%%%%%%%%
\begin{figure}[!t]
\centering
\includegraphics[scale=0.25]{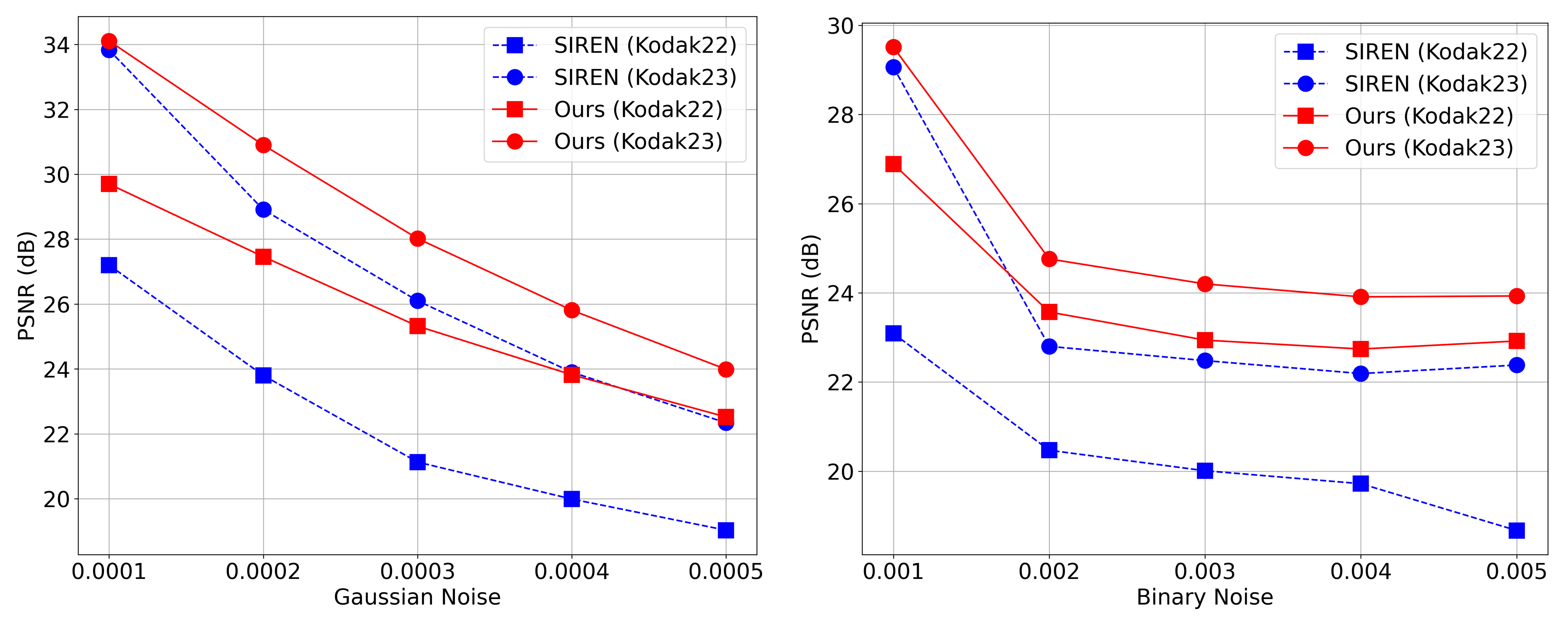}
\vspace{-5pt}
\caption{Comparison of baseline SIREN and our method on {\tt Kodak22} and {\tt Kodak23} images reconstruction under various Gaussian and binary noise conditions.} 
\label{fig:rgb_image_plot}
\vspace{-0.4cm}
\end{figure}
%%%%%%%%%%%%%%%%%%%%%%
Figure~\ref{fig:rgb_image} compares the reconstruction results of the vanilla model and our method. It can be seen that our method preserves more details under both Gaussian and binary noise. Moreover, we test different noise strengths of both types and report the PSNR in Figure~\ref{fig:rgb_image_plot}. Although there is performance degradation in noise-free conditions, our method demonstrates consistently superior performance compared to the vanilla model.

%%%%%%%%%%%%%%%%%%%%%%
\begin{figure}[!t]
\centering
\includegraphics[scale=0.3]{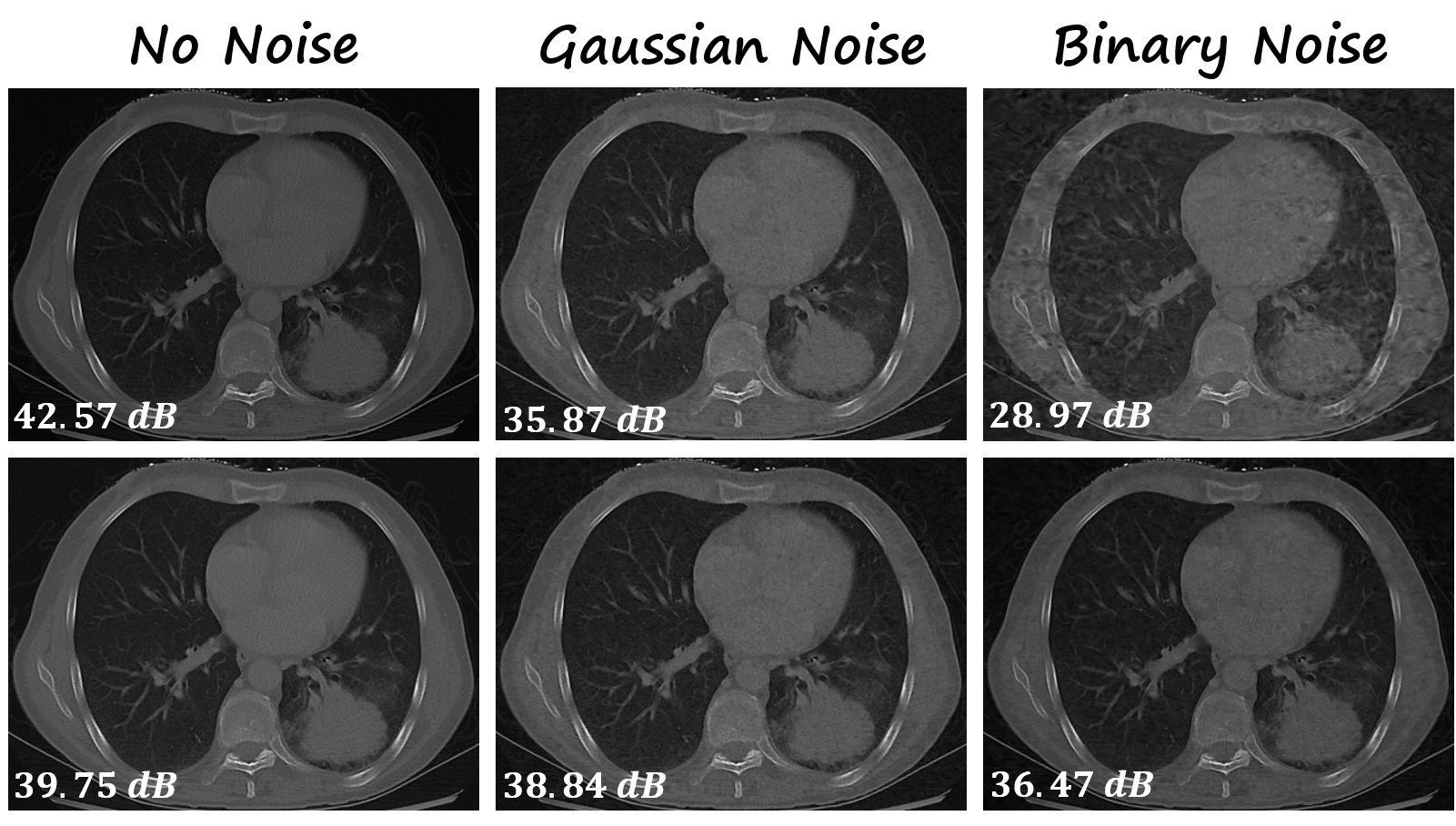}
% \vspace{-5pt}
\caption{Reconstruction comparison of baseline SIREN (first row) and our method (second row) for \textbf{CT image} reconstruction under Gaussian and binary noise conditions.} 
\label{fig:ct}
\vspace{-0.2cm}
\end{figure}
%%%%%%%%%%%%%%%%%%%%%%
\begin{figure}[!t]
\centering
\includegraphics[scale=0.28]{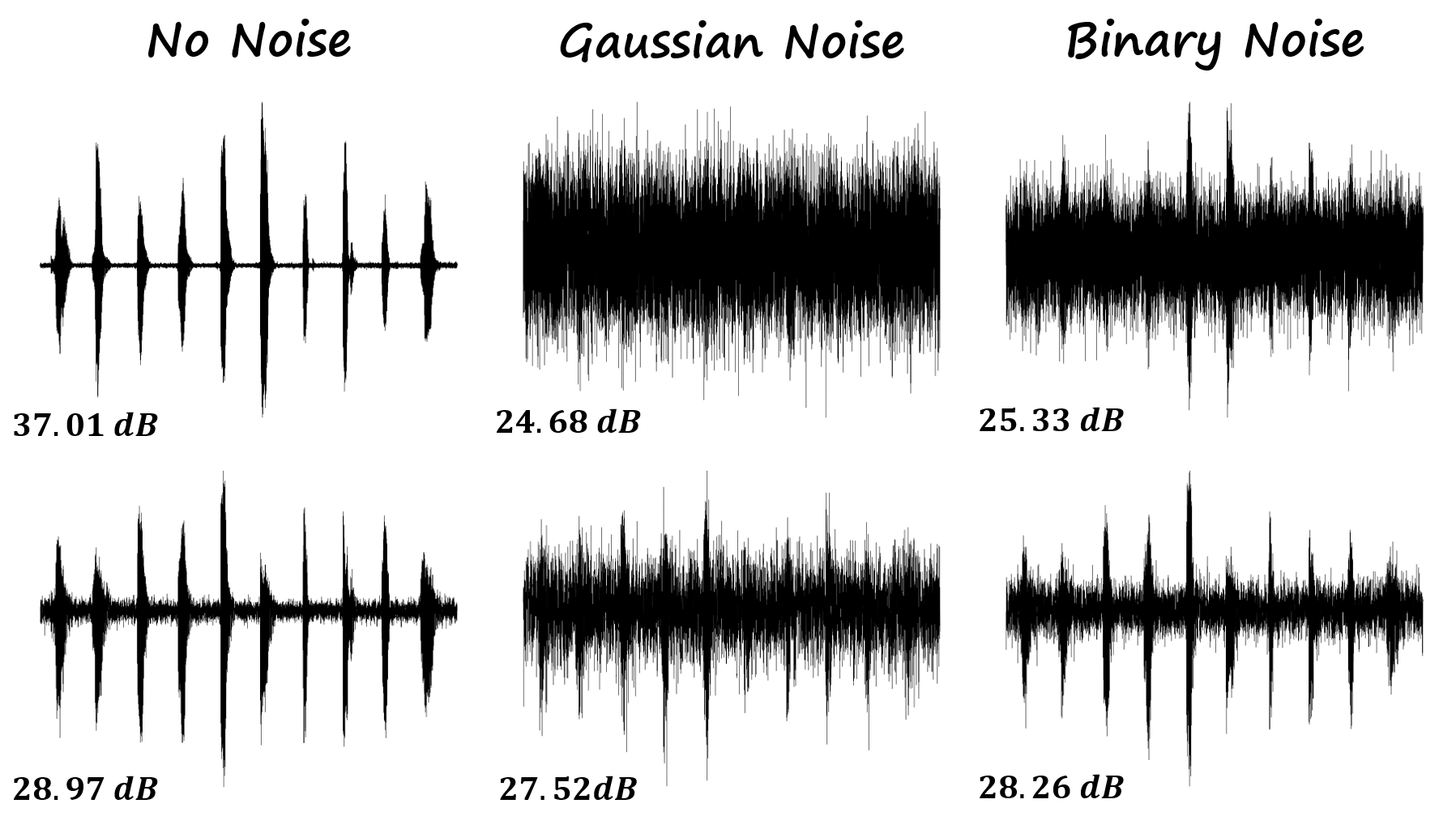}
% \vspace{-5pt}
\caption{Comparison of baseline SIREN (first row) and our method (second row) for \textbf{audio signal} reconstruction under Gaussian and binary noise conditions.} 
\label{fig:audio}
\vspace{-0.2cm}
\end{figure}
%%%%%%%%%%%%%%%%%%%%%%
\begin{figure}[!t]
\centering
\includegraphics[scale=0.4]{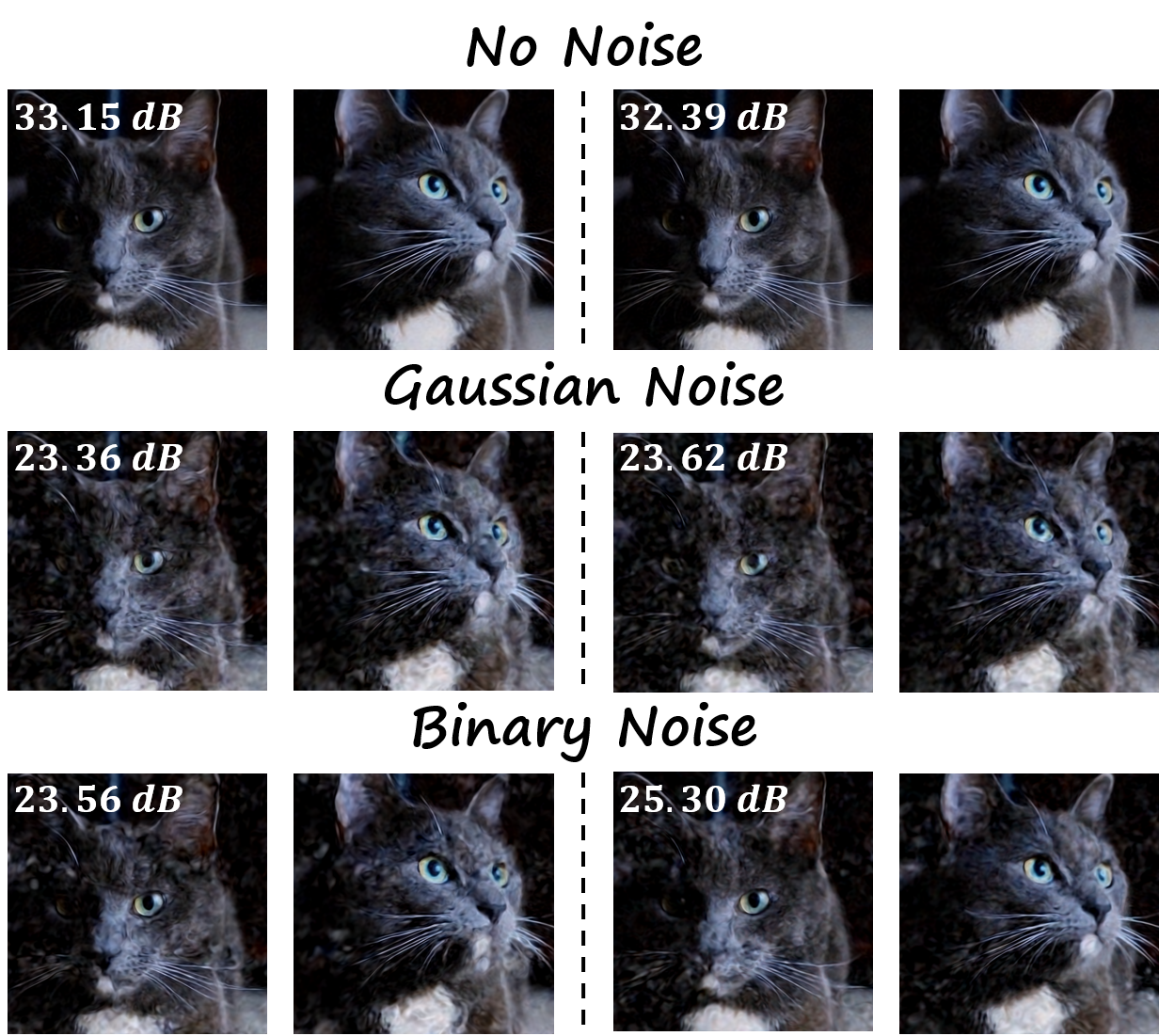}
% \vspace{-5pt}
\caption{Comparison of baseline SIREN (left two columns) and our method (right two columns) for \textbf{video signal} reconstruction under Gaussian and binary noise conditions.} 
\label{fig:video}
\vspace{-0.2cm}
\end{figure}
%%%%%%%%%%%%%%%%%%%%%%

\noindent \textbf{More modality results.} To demonstrate the generalization capability of our proposed method, we conduct more experiments across multiple modalities, including medical computed tomography (CT) image in Figure~\ref{fig:ct}, audio in Figure~\ref{fig:audio} and video in Figure~\ref{fig:video}~\cite{pistilli2022signal, molaei2023implicit}. In Figure~\ref{fig:ct}, the baseline SIREN outperforms our approach by 2.82 dB in the absence of noise, but our method are more robust under noise conditions by yielding up to a 7.5 dB in PSNR value. For audio signal reconstruction, a similar trend can also be observed: the reconstruction of our method is 8.04 dB lower than that of the vanilla SIREN model but achieves a PSNR increase of up to 2.84 dB under Gaussian noise and 2.93 dB under binary noise. For video reconstruction, as shown in Figure~\ref{fig:video}, we conduct an experiment on a 2-second video with 25 frames per second. As expected, our method excels in noisy environments, delivering a PSNR increase of up to 0.26 dB under Gaussian noise and 1.74 dB under binary noise. All the above experiments demonstrate that our method can effectively enhance the robustness of INRs in various applications.
%%%%%%%%%%%%%%%%%%%%%%
\begin{table}[!t]
\renewcommand{\arraystretch}{1.2}
\caption{Ablation study on the hyperparameter of $\lambda$ on {\tt Kodak} images under Gaussian noise ($\sigma=0.001$).}
\centering
\begin{tabular}{lcccc}
\toprule
\multirow{2}{*}{\textbf{Value of $\lambda$}} & \multicolumn{4}{c}{\textbf{PSNR (dB)}} 
\\
\cmidrule{2-5} 
& {\tt Kodak01} & {\tt Kodak08} & {\tt Kodak12} & {\tt Kodak24} \\ \hline
$\lambda = 0.01$                       & 16.20       & 13.42       & 19.89      & 15.24 \\
$\lambda = 0.1$                        & 17.64       & 14.48       & 19.61      & 16.63 \\
$\lambda = 0.2$                        & 17.61       & 14.43       & 20.21      & 16.53 \\
$\lambda = 0.5$                        & 18.17       & 14.05       & 19.34      & 16.71  \\
\bottomrule
\end{tabular}
\label{tab:ablation}
\vspace{-0.4cm}
\end{table}
%%%%%%%%%%%%%%%%%%%%%%

\subsection{Ablation Study}
The ablation study presented in Table \ref{tab:ablation} examines the impact of varying the regularization hyperparameter $\lambda$ on the performance of our method, as measured by PSNR across four Kodak images, which is the same as those in Table \ref{tab:result} for consistency under Gaussian noise ($\sigma = 1e^{-3}$). The results demonstrate that reconstruction quality is highly dependent on the choice of $\lambda$. Generally, a small $\lambda$ value of 0.01 leads to sub-optimal PSNR values, with \texttt{Kodak08} showing the lowest PSNR of 13.42 dB. As $\lambda$ increases to 0.1, the PSNR improves across all images, with \texttt{Kodak01} achieving the highest PSNR of 17.64 dB. Further increasing $\lambda$ to 0.2 results in continued improvements for some images, particularly \texttt{Kodak12}, which reaches a PSNR of 20.21 dB. Even a larger value of $\lambda$, such as 0.5, can lead to optimal performance in some cases, like \texttt{Kodak01} and \texttt{Kodak24}, which present the highest PSNR of 18.17 dB and 16.71 dB.

%%%%%%%%%%%%%%%%%%%%%%%%%%%%%%%%%%%%%%
%             Conclusion             %
%%%%%%%%%%%%%%%%%%%%%%%%%%%%%%%%%%%%%%
\section{Conclusion}
\label{sec:conclusion}
In this paper, we systematically investigated the vulnerabilities of INRs to weight perturbations for the first time. To address this issue, we formulate the robustness problem and derive a gradient-based loss to improve the model performance under noisy conditions. Extensive experiments on reconstruction tasks on multiple modalities, including natural image, CT, audio, and video, demonstrate that our method significantly enhances model robustness under various noise conditions by achieving gains of up to 7.5 dB in PSNR value compared to the baseline approach.

\vfill\pagebreak

\label{sec:refs}
\bibliographystyle{IEEEbib}
\bibliography{refs}

\end{document}